\documentclass{article}

\usepackage{amsmath,epsfig}
\usepackage[preprint]{spconfa4}

\copyrightnotice{\textbf{978-1-6654-3864-3/21/\$31.00~\copyright 2021 IEEE}}

\let\OLDthebibliography\thebibliography
\renewcommand\thebibliography[1]{
  \OLDthebibliography{#1}
  \setlength{\parskip}{0pt}
  \setlength{\itemsep}{0pt plus 0.3ex}
}

\usepackage{times}
\usepackage{epsfig}
\usepackage{graphicx}
\usepackage{amsmath}
\usepackage{amssymb}
\usepackage{caption}

\usepackage{booktabs}
\usepackage{multirow}
\usepackage{enumitem}
\usepackage{dsfont}
\usepackage{balance}
\usepackage{placeins}
\usepackage{stfloats}

\usepackage{hyperref}
\hypersetup{
    colorlinks=true,
    urlcolor=blue,
}
\begin{document}\sloppy

\title{DeepWORD: A GCN-based Approach for Owner-member Relationship Detection in Autonomous Driving}
%
\name{Zizhang Wu$^{1,\ast}$, Man Wang$^{1}$, Jason Wang$^{1}$, Wenkai Zhang$^{1}$, Muqing Fang$^{2}$, Tianhao Xu$^{3}$}
\address{$^{1}$Zongmu Technology; $^{2}$Politecnico di Torino; $^{3}$Technical University of Braunschweig\\
{\tt \small zizhang.wu@zongmutech.com}}

\maketitle


\begin{abstract}
 It's worth noting that the owner-member relationship between wheels and vehicles has an significant contribution to the 3D perception of vehicles, especially in the embedded environment. However, there are currently two main challenges about the above relationship prediction: i) The traditional heuristic methods based on IoU can hardly deal with the traffic jam scenarios for the occlusion. ii) It is difficult to establish an efficient applicable solution for the vehicle-mounted system. To address these issues, we propose an innovative relationship prediction method, namely \textbf{DeepWORD}, by designing a graph convolution network (GCN). Specifically, we utilize the feature maps with local correlation as the input of nodes to improve the information richness. Besides, we introduce the graph attention network (GAT) to dynamically amend the prior estimation deviation. Furthermore, we establish an annotated owner-member relationship dataset called \textbf{WORD} as a large-scale benchmark, which will be available soon. The experiments demonstrate that our solution achieves state-of-the-art accuracy and real-time in practice. The \textbf{WORD} dataset will be made publicly available at \href{https://github.com/NamespaceMain/owner-member-relationship-dataset}{https://github.com/NamespaceMain/owner-member-relationship-dataset}.

\end{abstract}


\section{INTRODUCTION}
\begin{figure}[!t]
    \centering
    \includegraphics[width=7.5cm]{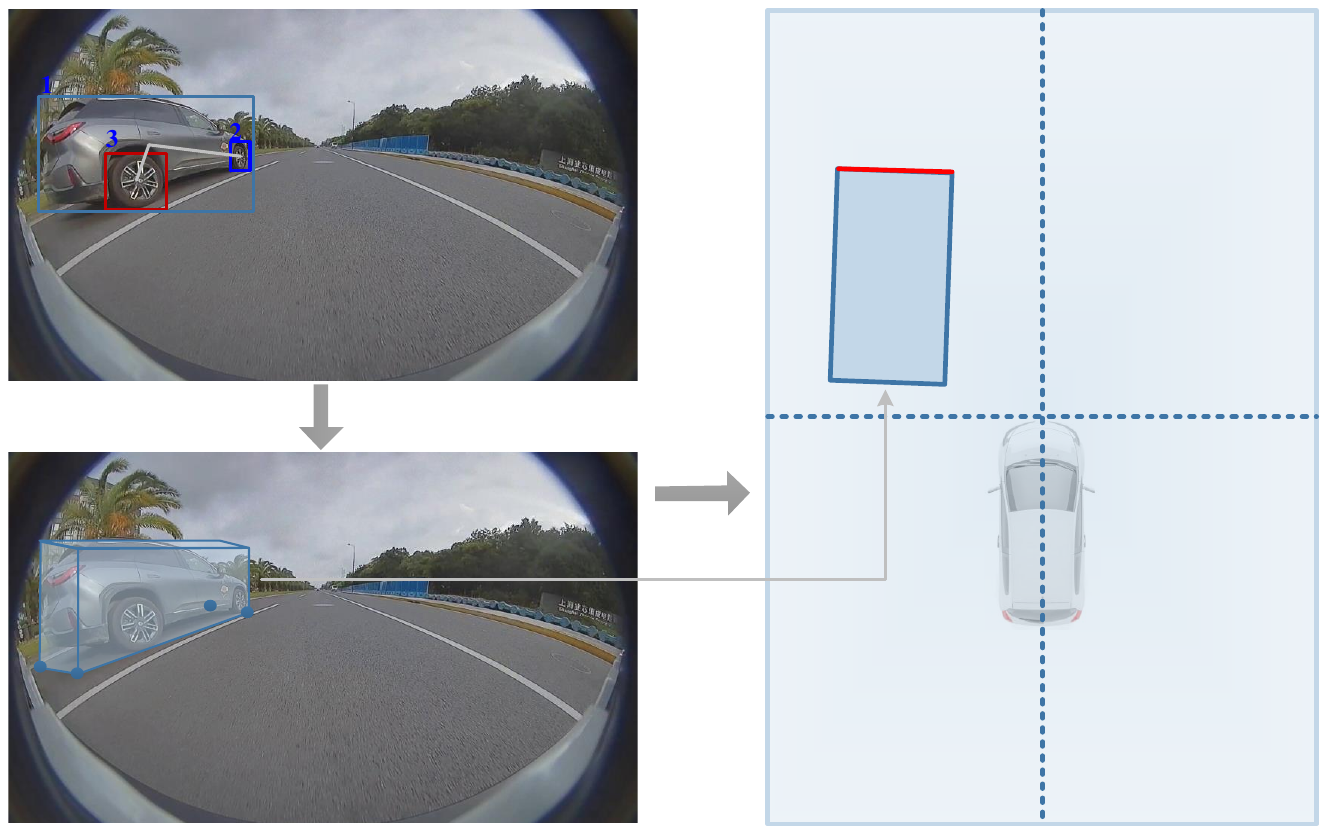}
    \caption{The visualization of the owner-member relationship between wheels and vehicles on 3D information acquisition. The estimation of four BEV corners come into being after obtaining the relationship between wheels and vehicles. In this way, the projection position of surrounding vehicles is helpful for the autonomous driving system to make decision.}
    \label{fig:Figure01}
\end{figure}

\begin{figure}[!t]
    \centering
    \includegraphics[width=8cm]{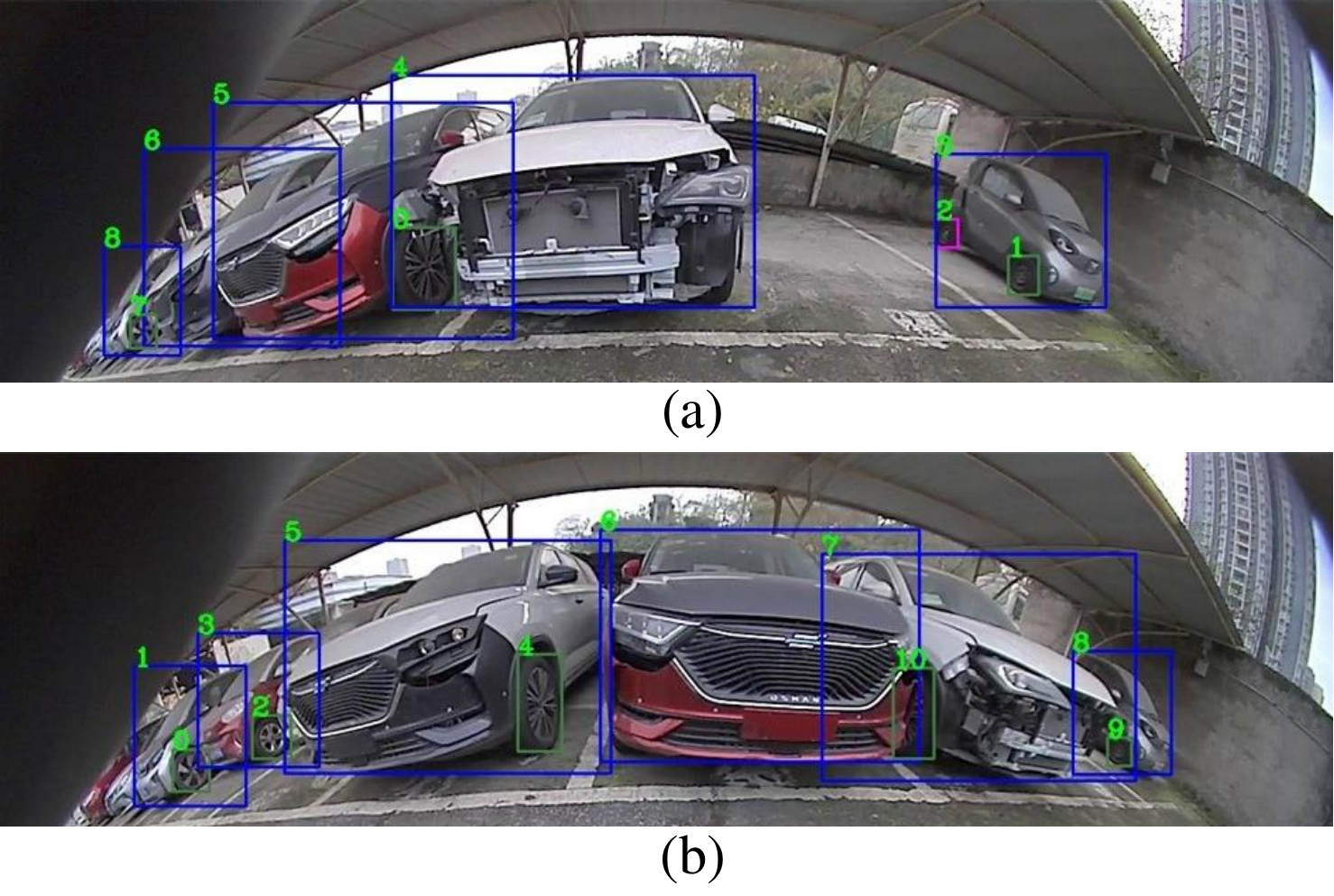}
    \caption{The visualization of complex scenarios. It is obvious that the bounding-boxes of vehicles 4 and 5 completely contain wheel 0 as shown in (a), and the bounding-boxes of vehicles 7 and 8 also completely contain wheel 9 as shown in (b). In these scenarios, the prediction of the owner-member relationship is not available with the IoU alone.}
    \label{fig:Figure02}
\end{figure}

With the rapid development of autonomous driving \cite{article13}, the 3D perception has drawn increasing attention and shown great potential in the vehicle-mounted system \cite{article14, article15}, which is critical to localization \cite{article16,  article17}, planning, and obstacle avoidance \cite{article18, article19}, etc., as shown in Figure \ref{fig:Figure01}. Among existing methods \cite{article20, article21, article22}, they usually employ the homography transformation \cite{article20} of the wheel grounding points obtained from the wheel's detection results to estimate the localization of target vehicles with limited computing ability in the low cost environment. Consequently, it is necessary to introduce the owner-member relationship of wheels and vehicles to constrain the process. While researchers usually employ IoU to predict relationships of different objects, due to the difficulty of embedded network transplantation. It is inappropriate to predict the relationship solely through IoU for complex scenarios, as shown in Figure \ref{fig:Figure02}, in which two different vehicle bounding-boxes completely contain the same wheel.

In this paper, in order to cope with various scenarios and get rid of the constraints of prior analysis, we propose a novel owner-member relationship prediction method by introducing a graph convolution network (GCN) to implicitly learn the relationship between the wheels and vehicles.

\section{RELATED WORK AND OUR CONTRIBUTIONS}

\subsection{Related work}
\textbf{Traditional relationship prediction methods.}
Traditional relationship prediction methods generally observe the statistics and set constraint ranges by the attributes of objects, such as semantic dependency, pose, and order constraint, and then design the appropriate models for training and prediction. \cite{article3} proposes a model based on the spatial layout information of the detection object to predict the relationship. \cite{article4} employs the CRF (Conditional Random Field) \cite{article25} to obtain statistics prior of owner-member relationship between objects and builds a model to predict the correlation between objects. \cite{article5} regards the common occurrence frequency of object pairs as prior knowledge and utilizes LSTM (Long short-term memory) \cite{article26} as an encoder to transfer context information to improve the feature representation between objects.

\textbf{Graph convolutional network methods.}
The frequency domain based GCN standard mathematical expression paradigm is constructed after the exploration and optimization of series methods \cite{article6, article7, article8, article9}, so that the graph convolutional structure can be trained like the general convolutional structure. Specially, GCN-based relationship prediction methods are mainly manifested in the prediction of the owner-member and relative location relationships between objects. \cite{article10} proposes the graphSAGE method, which expounds the understanding of the aggregation and updates operator of graph neural networks from the perspective of the space domain. It also summarizes the mathematical expression in the space domain and explores the optimization method of the training problem of large-scale graph data structure. Moreover, to better predict the importance of the connection relationship between nodes, \cite{article11} introduces the attention mechanism into the graph neural network. Besides, \cite{article12} tries to encode the relationship through a graph neural network and introduces a spatial-aware graph relationship network (SGRN) for automatic discovery. It combines key semantics and spatial relationships at the same time to achieve better relationship prediction.


\subsection{Our motivations and contributions}
After the exploration of the above methods, we find that there are still some challenges in the owner-member relationship prediction of wheels and vehicles.

Firstly, there is no available public dataset for this task in the field of autonomous driving, which brings huge difficulties to deep learning based methods. Secondly, the above methods usually predict the owner-member relationship of wheels and vehicles with the prior statistical distribution, such as the IoU or the relative position of wheels and vehicles. Thus, they are incapable of covering all the scenarios, which is unbeneficial to the generalization of the method. Moreover, it is difficult to build an effective model by stacking a series of logical judgments for complicated scenarios.

In this work, we attempt to fill the research gaps to some extent. So we propose a GCN based owner-member relationship prediction method for wheels and vehicles. Specifically, to take full advantage of the prior knowledge, we make an analysis of geometrical relative location statistics between wheels and vehicles and consider them as prior parameters. In addition, to more efficiently supervise the learning process, we introduce the GCN structure to modeling the owner-member relationship between the wheels and vehicles. Furthermore, we utilize the feature vectors with local correlation as the input of nodes for improving the information richness of the nodes in GCN. Finally, to decrease the negative impact of noise in prior statistical data, the GAT \cite{article11} module is introduced to dynamically amend the prior estimation deviation of edges through the training process. Our contributions are summarized as follows:

1) We establish a large-scale benchmark dataset WORD including 9,000 samples, which is the known first available relationship dataset in the field of autonomous driving and will be public as soon.

2) We propose a GCN-based owner-member relationship prediction network for wheels and vehicles, which can excellently cope with the relationship prediction in different complicated scenarios.

3) We validate the effectiveness of our proposed method through the WORD dataset. The experiment results show that this method achieves superior accuracy, and especially real-time effects in an embedded environment.


\section{DEEPWORD: A GCN-BASED APPROACH FOR OWNER-MEMBER RELATIONSHIP DETECTION}
\begin{figure*}[!t]
    \centering
    \includegraphics[width=17cm]{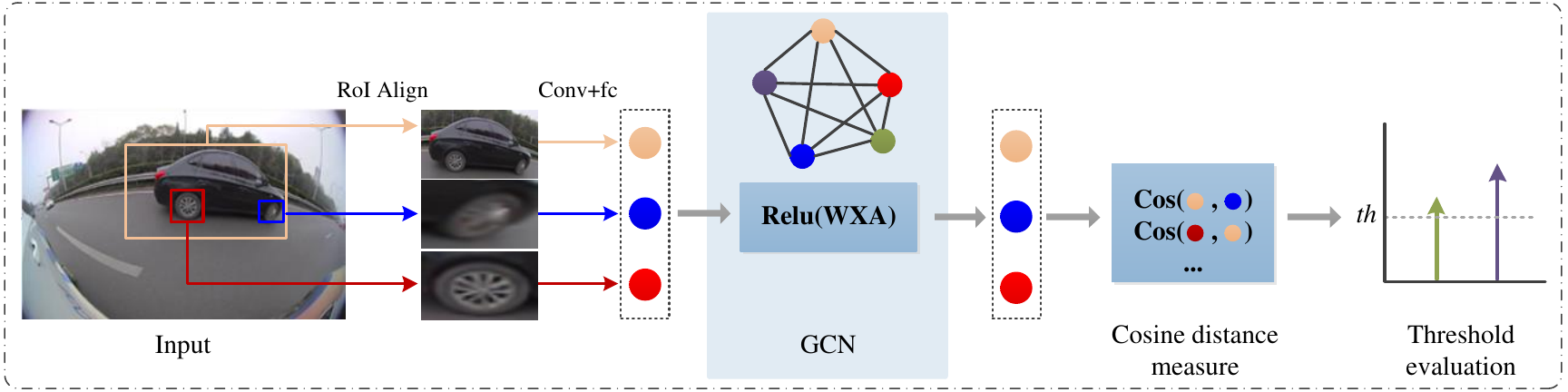}
    \caption{The overall framework of the proposed DeepWORD. The input is the detected bounding box, and after ROI Align the images are the same size. Whereafter, it generates corresponding feature vectors with MLP operation, learning the owner-member relationship with GCN to update the feature vectors. Further, we calculate the cosine distance between the feature vectors from the wheels and vehicles, and retain the wheel-vehicle pairs greater than the threshold as the final results.}
    \label{fig:Figure03}
\end{figure*}

		
			
		

In this section, we introduce the proposed method in detail, which is about the owner-member relationship between the wheels and vehicles based on the graph convolution. 

\subsection{The overall framework of proposed method}

The overall framework of the proposed DeepWORD is shown in Figure \ref{fig:Figure03}, which consists of a detection network and a GCN-based owner-member relationship prediction network. To achieve more efficient detection, we employ CenterNet \cite{article24} as the object detection network and feed the detection results into the relationship prediction network to predict the owner-member relationship between the wheels and vehicles.

Specially, we use two Gaussian mixture distributions from prior statistics as the initial value of the edge in the relationship prediction network. Moreover, in order to amend the deviation from prior statistics, we introduce the GAT into GCN to dynamically update edges. The updated feature vectors of each object are available after GCN, which can fuse more global semantic information to improve the similarity of the wheels and vehicles that belong to the identical vehicle. Subsequently, cosine distance can measure the similarity of the wheels and vehicles and retain the wheel-vehicle pairs with scores greater than the default threshold 0.5.

\subsection{GCN-based relationship prediction network}

In this section, we cover the implementation details of the proposed GCN-based relationship prediction network.



\begin{figure}[!h]
    \centering
    \includegraphics[width=5cm]{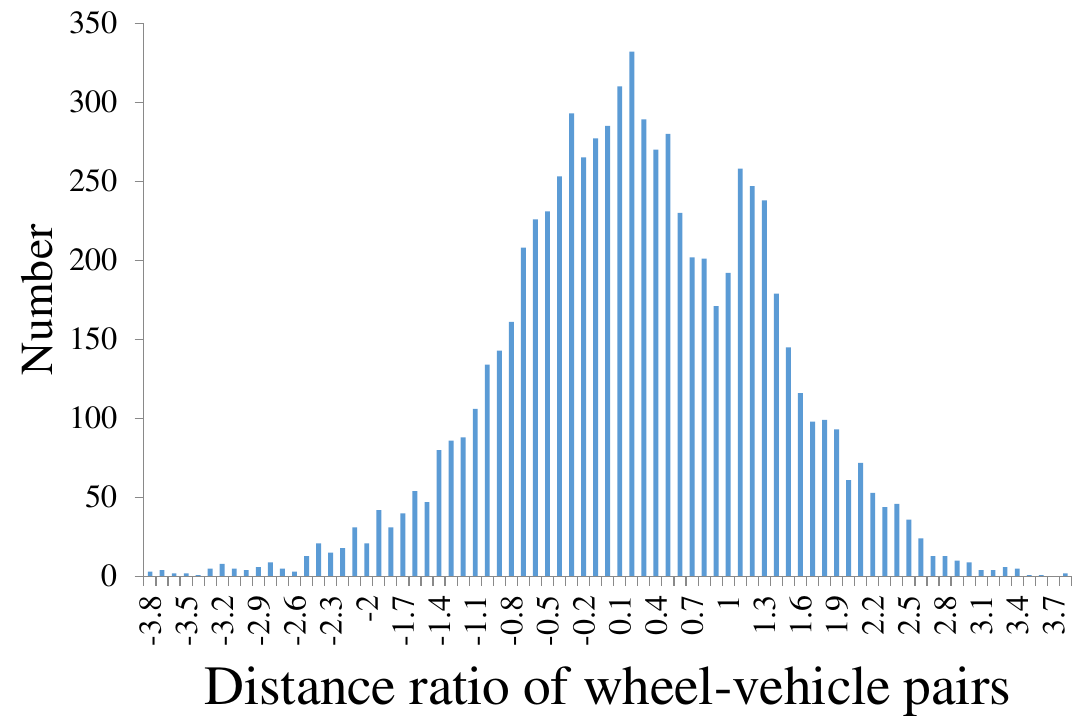}
    \caption{The distance ratio distribution of wheel-vehicle pairs.}
    \label{fig:Figure06}
\end{figure}

\textbf{Prior statistical relationship.} The prior statistical relationship between wheels and vehicles is helpful to the update of nodes in GCN. In our method, we employ the Gaussian mixture distribution to model the position relationship between wheels and vehicles, as shown in Figure \ref{fig:Figure06}. We find that there are identical relationships of the bounding-boxes of wheels and vehicles that belong to the identical vehicle: For most bounding-boxes of vehicles, the distances with its subordinate wheels are shorter than that with other wheels. Meanwhile, the most bounding-boxes of wheels generally locate in the bottom half location of the vehicle bounding-boxes. 

As stated above, we conduct a statistical analysis of the distance ratio of wheel-wheel and wheel-vehicle pairs. In detail, to eliminate the interference of different sizes of objects, we normalize them as follows:
\setlength\abovedisplayskip{5pt}
\setlength\belowdisplayskip{5pt}
\begin{equation}
D^2  = (\frac{A_\text{j}}{W}-\frac{B_\text{j}}{W})^2+(\frac{A_\text{i}}{H}-\frac{B_\text{i}}{H})^2
\label{Eq(1)}
\end{equation}
\noindent
where $A$ means the vehicle with $B$ means the wheel in the wheel-vehicle pair, while in the wheel-wheel pair, $A$ is the rear wheel and $B$ is the front wheel. $A_\text{j}$ and $B_\text{j}$ represent the horizontal positions of $A$ and $B$, $A_\text{i}$ and $B_\text{i}$ refer to the vertical positions of $A$ and $B$. $W$ and $H$ are the width and height of the input image.


To relieve the influence of image distortion and the scale changes from the near to the distant, we employ Eq.\ref{Eq(2)} to obtain the distance ratio between vehicles and wheels. After the log transformation, it is obvious that the distribution accords with the Gaussian mixture model. We also apply this data processing approach to the front-rear wheel pairs. Consequently, with two Gaussian mixture models it models the above two distance ratios separately, which can work as the prior statistic of adjacent matrices in GCN.
\setlength\abovedisplayskip{5pt}
\setlength\belowdisplayskip{5pt}
\begin{equation}
Ratio = \frac{2D}{W_\text{B}+H_\text{B}}
\label{Eq(2)}
\end{equation}
\noindent
where $W_\text{B}$ and $H_\text{B}$ represent the width and height of $B$.







\textbf{GCN structure.} In the proposed GCN structure, we utilize the feature vectors obtained by convolutional layers plus fully connected layers (\textbf{Conv+fc}) as nodes, which makes our model tend strongly to express local spatial information. It is noteworthy that we employ the above-mentioned prior statistic relationship to initialize the corresponding values of the adjacency matrix. Finally, we calculate the cosine distance of the updated nodes information of vehicle-wheel pairs after the GCN structure and then retain the pairs whose scores are greater than the default threshold as the final result pairs. Here we take the threshold as 0.5 after experiments.

\textbf{GAT module.} Given that the prior statistics are greatly related to the number of samples and the scene richness, we introduce the GAT into GCN to amend the deviation caused by the limited data. Specifically, we can weight GAT linearly for each edge and secondly refine the weights of the edges in GCN, which can alleviate the impact of noise in the available dataset and enhance the representation ability of the network. 

We set the nodes vector as $\textbf{h}=\left \{ {\overrightarrow{h}_{\text{1}}},{\overrightarrow{h}_{\text{2}}},...,{\overrightarrow{h}_{\text{N}}}\right \}$, where ${\overrightarrow{h}_{\text{i}}}\in R^{F}$, $N$ represents the number of nodes, and $F$ is the number of features in each node. In GCN structure, we input the features of each node $\overrightarrow{h}$ and their adjacent nodes $\overrightarrow{h}_{\text{i}}$ into the GAT module, and concatenate them through two fully connected layers. Then we extract the weight matrix $\textbf{W}= \left \{ w_{\text{1}},w_{\text{2}},...,w_{N\times N} \right \}$ by fully-connected layers (FC) and nonlinear activation layers, where $w_{\text{i}}\in R^{F\times F}$.
\setlength\abovedisplayskip{5pt}
\setlength\belowdisplayskip{5pt}
\begin{equation}
Net(X_\text{1},X_\text{2}) \rightarrow FC(Relu(FC(Concat(X_\text{1},X_\text{2}))))
\label{Eq(3)}
\end{equation}
\noindent
where $Net$ denotes the procedure of dealing with GCN, $(X_{\text{1}},X_{\text{2}})$ represents the features of each node with their adjacent nodes, $FC$ is a fully connected layer.

Following, we utilize the softmax function to normalize the output and obtain the attention coefficient $Scale_i$:
\setlength\abovedisplayskip{5pt}
\setlength\belowdisplayskip{5pt}
\begin{equation}
Scale_{\text{i}}=\frac{exp(Net(\overrightarrow{h},\overrightarrow{h}_{\text{i}}))}{\sum _{\text{j}}exp(Net(\overrightarrow{h},\overrightarrow{h}_{\text{i}}))}
\label{Eq(4)}
\end{equation}
Finally, we update the original edge weight $w_{\text{i}}$ between the node and the adjacent node by multiplying it with $Scale_{\text{i}}$ to get a new weight $w_{\text{i}}^{'}$ as follows:
\setlength\abovedisplayskip{5pt}
\setlength\belowdisplayskip{5pt}
\begin{equation}
w_{\text{i}}^{'} = w_{\text{i}}\times Scale_{\text{i}}
\label{Eq(5)}
\end{equation}

In this way, we can dynamically correct the prior estimation deviation of the edge during the training process, and improve the prediction accuracy of the owner-member relationship between wheels and vehicles.

\textbf{Label preparation.} To obtain appropriate label for relationship prediction network, firstly, we resize the obtained feature maps of the wheels and vehicles to $H\times W$ and normalize each value to [0, 1]. Hence, four normalized coordinates of objects constitute a matrix of $H\times W$. The concatenation of the coordinate matrix and the matrix of wheels and vehicles serves as the input feature of the relationship network for $H\times W\times \text{7}$. Besides, we build a Gaussian mixture distribution to model the available data. Then we calculate the probability of wheel-vehicle pairs and wheel-wheel pairs to generate the initial adjacency matrix. Moreover, we also set the values of unnecessary objects to 0 at corresponding positions in the adjacent matrix, such as small objects.

\section{EXPERIMETS}
\subsection{Dataset overview}
We construct a dataset called WORD (Wheel and Vehicle Owner-Member Relationship Dataset) as a benchmark, which contains two typical categories in autonomous driving: parking scene and highway scene. The WORD contains about 9,000 images, which are collected by a surround-view camera system composed of 4 fisheye cameras. In the WORD, the images with same frame id represent the recording of the environment from different camera's view simultaneously.
\begin{figure*}[!t]
    \centering
    \includegraphics[width=17cm]{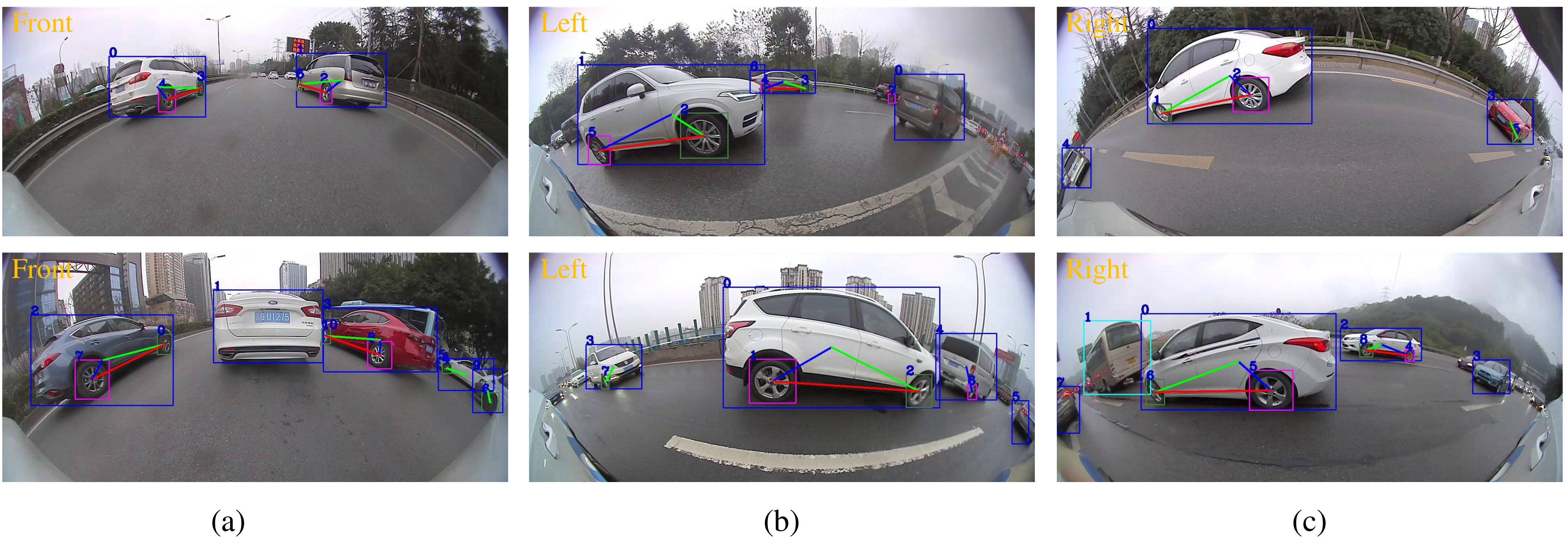}
    \caption{The visualization of the owner-member relationship between wheels and vehicles. (a) Front camera; (b) Left camera; (c) Right camera. Where the upper left corner  is the number of each object, the red line between two wheels means that they are a couple. The green and blue lines respectively connect the rear and front wheels and the vehicles that they belong to.}
    \label{fig:Figure04}
\end{figure*}


\subsection{Improvements on embedded platforms}

In order to deploy our model to the embedded platform of autonomous driving system, we deeply analyze the Qualcomm SNPE acceleration library and find that some operations in the proposed model are not supported. These operations mainly include ROI Align, GCN matrix multiplication operations, multiple heads and sizes input, etc. Therefore, we make some improvements in the proposed model as follows:
 
(1) we remove the RoI Align and resize the feature map extracted from the original image to a fixed size ($\text{56} \times \text{56}$).
 
(2) we use a fully connected layer instead of the matrix multiplication operation in GCN.
 
(3) To solve the problems deriving from the multiple heads and sizes input, we remove the FC encoding module of the coordinate value and concatenate it with the cropped vehicles and wheels from the original image to generate the input features in size of $H \times W \times \text{7}$.






\subsection{Backbone selection}

In order to deploy our model to the embedded platform of autonomous driving, we conduct a series of experiments in backbone selection as shown in Table \ref{table01}. We select the structure of Conv+fc as our backbone to extract feature maps after considering the trade-off of speed and accuracy.

\begin{table}[!h]
\setlength{\tabcolsep}{5pt}
\centering
\small
\caption{Performance comparison of backbone selection. $AP_{v}$ represents the accuracy of image visualization. 10-speed and 1-speed indicate the speed at which the model processes the image when the batch size is 10 and 1, respectively. }
\begin{tabular}{@{}lcccc@{}}
\toprule
Backbone & \begin{tabular}[c]{@{}c@{}}Model size\\ (M)\end{tabular} & \begin{tabular}[c]{@{}c@{}}$AP_{v}$\\ (\%)\end{tabular} & \begin{tabular}[c]{@{}c@{}}10-Speed\\ (ms/10imgs)\end{tabular} &
\begin{tabular}[c]{@{}c@{}}1-Speed\\ (ms/img)\end{tabular}  \\ \midrule
Conv+fc       & 68                                                      & 62.83                                                 & 28                                                            & 5       \\
ResNet18 & 66                                                     & 92.47                                                 & 110                                                           & 13      \\ \bottomrule
\end{tabular}

\label{table01}
\end{table}
 
Furthermore, we also conduct experiments to find the equilibrium between speed and accuracy after confirming the backbone, as shown in Table \ref{table02}. The results guide us when appropriately deepen the convolutional layers and reduce the amount of fully connected layers, we reduce the size of the model to 28M. The model achieves the best performance of 95.70\% on image visualization accuracy, and its running speed is almost double faster than Conv+fc.
\begin{table}[!h]
\setlength{\tabcolsep}{3pt}
\small
\caption{Performance comparison of different parameters setting based on Conv+fc. The measures taken in turn for backbone are: 1) Neg0.1: reduces the weight of negative samples to 0.1; 2) -Neg: decreases the number of negative samples; 3) -Sma: uses a mask to filter too small objects; 4) 56: intercepts the input image size to 56; 5) 56\_ex\_4: deepens 4 convolutions and reduce the amount of fully connected layers.}
\begin{tabular}{@{}lcccc@{}}

\toprule
Backbone            & \begin{tabular}[c]{@{}c@{}}Model size\\ (M)\end{tabular} & \begin{tabular}[c]{@{}c@{}}$AP_{v}$\\ (\%)\end{tabular} & \begin{tabular}[c]{@{}c@{}}10-Speed\\ (ms/10imgs)\end{tabular} &
\begin{tabular}[c]{@{}c@{}}1-Speed\\ (ms/img)\end{tabular} \\ \midrule
Conv+fc          & 68                                                       & 62.83                                                 & 28                                                            & 5       \\
Conv+fc Neg0.1          & 68                                                       & 69.34                                                 & 28                                                            & 5       \\
Conv+fc-Neg           & 68                                                       & 73.42                                                 & 28                                                            & 5       \\
Conv+fc-sma           & 68                                                       & 89.21                                                 & 28                                                            & 5       \\
Conv+fc 56           & 53                                                       & 90.33                                                 & 18                                                            & 4       \\
Conv+fc 56\_ex\_4 & \textbf{28}                                                       & \textbf{95.70}                                & \textbf{13.1}                                          & \textbf{3.2}     \\ \bottomrule
\end{tabular}

\label{table02}
\end{table}

\subsection{Training of the relationship network} 

At the training stage, the structure of Conv+fc extracts the features of the input feature matrix. Firstly, the input matrix goes through Conv+fc, and GCN structure updates these feature vectors. Besides, we normalize the obtained features and calculate the cosine distance of each wheel and vehicle as the final relationship prediction results. Moreover, we use the L2 loss and adjust the weights of positive and negative samples manually to optimize the model. At the prediction stage, we multiply the predicted matrix with the mask, the position where the value is greater than 0.5 indicates that the corresponding combination has the owner-member relationship. The mask is mainly used to filter unnecessary objects. The visualization of the owner-member relationship between wheels and vehicles in highway scenes is shown in Figure \ref{fig:Figure04}.


\subsection{Performance evaluation of DeepWORD}

\begin{figure}[!h]
    \centering
    \includegraphics[width=8.5cm]{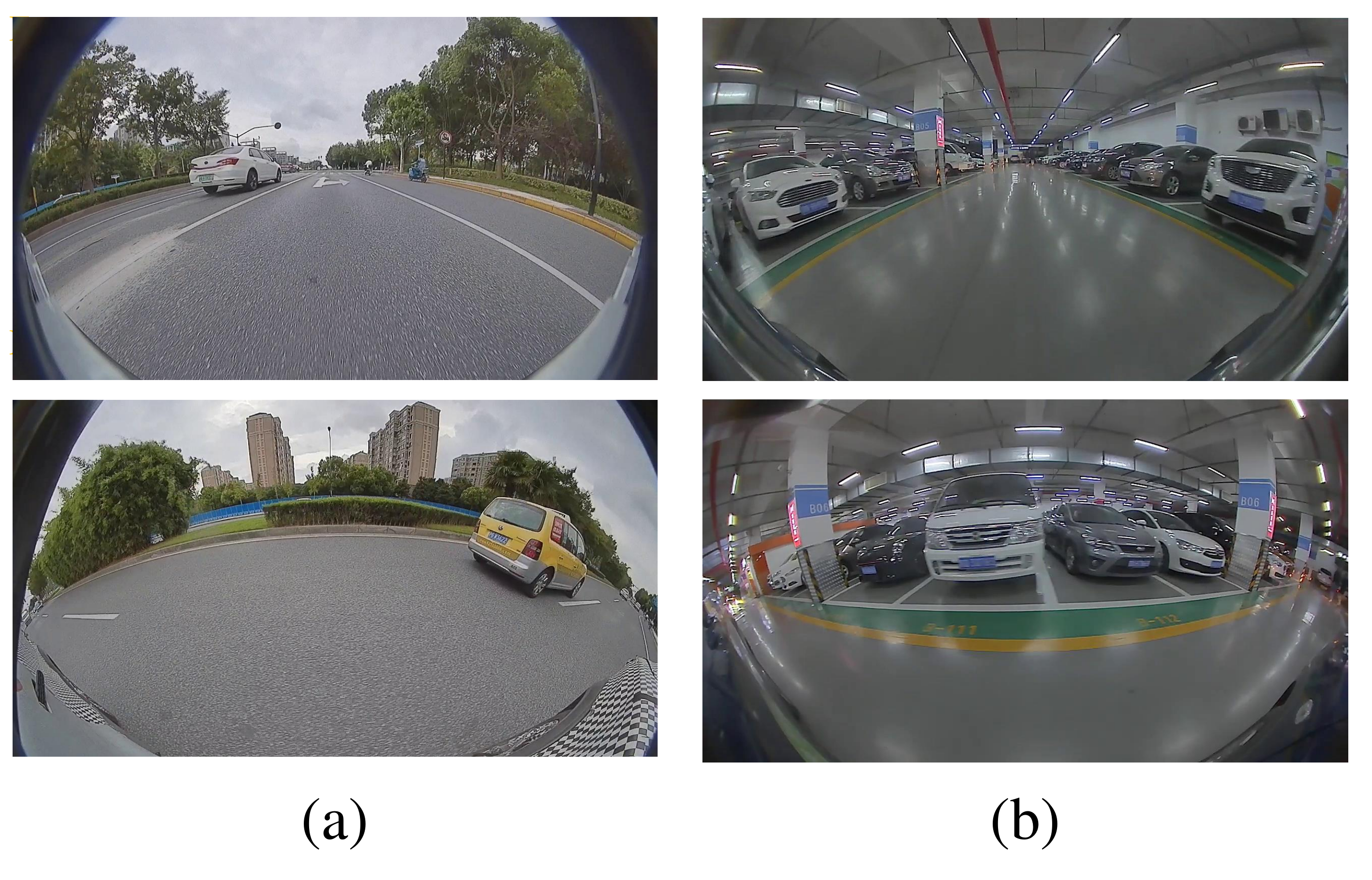}
    \caption{(a) and (b) are easy and hard samples of WORD.}
    \label{fig:Figure05}
\end{figure}

To prove the effectiveness of our structure, we compare the proposed DeepWORD with the previous logic model method. We select 1000 images to form an easy scene dataset in which each image contains no more than three vehicles, 1000 images to form a hard scene dataset where each image has more than three vehicles, and a mixed dataset consisting 500 easy images and 500 hard images. All of the samples in these three dataset are randomly sampled from the WORD. Some samples are shown in Figure \ref{fig:Figure05}. 

\begin{table}[!b]
\setlength{\tabcolsep}{15pt}
\centering
\small
\caption{Performance comparison of DeepWORD and logic model method.}
\label{table03}

\begin{tabular}{@{}lccc@{}}
\toprule
Methods             & Easy  & Hard  & Mixed \\ \midrule
Logic model & 92.91 & 71.83 & 79.90 \\
DeepWORD           &  \textbf{99.17} &  \textbf{94.35} &  \textbf{95.14} \\ \bottomrule
\end{tabular}
\end{table}

Furthermore, we perform experiments on these three datasets to compare classification accuracy. As shown in Table \ref{table03}, in an easy scene, our DeepWORD has a suspicious improvement by 6.26\% in terms of accuracy, compared with the logic model method. On hard and mixed datasets, the proposed DeepWORD has a significant improvement over the logic model method, and the accuracy has increased by 22.52\% and 15.24\% on each dataset. We check the results and find that the logic model method is likely to misjudge the owner-member relationship in the scene of dense vehicles parked vertically side by side. It is noteworthy that the DeepWORD is simpler and more efficient than the logic model method in terms of the scenario coverage capability and the accuracy of prediction.

\section{CONCLUSION AND FUTURE WORK}

In this paper, we propose a GCN-based owner-member relationship prediction network DeepWORD, which has good applicability and forecast precision in different scenarios. In particular, extensive experiments prove the efficiency and effectiveness of the proposed method in vehicle-mounted surround-view camera system. Besides, we establish and release a large-scale relationship dataset WORD. It is the first available dataset for relationship detection in the field of autonomous driving and is conducive to promote relevant research. In the future, we will continuously enlarge the WORD to include more real-world samples and will try to refine our owner-member relationship prediction solution DeepWORD.


\small
\bibliographystyle{IEEEbib}
\bibliography{icme2021template}

\end{document}